\documentclass[runningheads]{llncs}
\usepackage{graphicx}
\usepackage{url}            
\usepackage{booktabs}       
\usepackage{amsfonts}       
\usepackage{nicefrac}       
\usepackage{microtype}      
\usepackage{tikz}
\usepackage[misc]{ifsym}
\usepackage{tabularx}
\usepackage{hhline}
\usepackage{adjustbox}
\usepackage{multirow}
\usepackage{mathtools}
\usepackage{bbm}
\usepackage{color, colortbl}
\usepackage{algorithm,algorithmic}

\usepackage{cellspace}
\DeclarePairedDelimiterX\set[1]\{\}{\nonscript\,#1\nonscript\,}

\usetikzlibrary{shapes.geometric, arrows}
\tikzstyle{startstop} = [rectangle, rounded corners, minimum width=1.5cm, minimum height=0.8cm,text centered, draw=black, fill=red!30]
\tikzstyle{io} = [trapezium, trapezium left angle=70, trapezium right angle=110, minimum width=1.5cm, minimum height=0.8cm, text centered, draw=black, fill=blue!30]
\tikzstyle{process} = [rectangle, minimum width=1.5cm, minimum height=0.8cm, text centered, draw=black, fill=orange!30]
\tikzstyle{decision} = [diamond, minimum width=1.5cm, minimum height=0.8cm, text centered, draw=black, fill=green!30]
\tikzstyle{block} = [rectangle, rounded corners, minimum width=1.5cm, minimum height=0.8cm,text centered, draw=black, fill=orange!30]
\tikzstyle{network} = [rectangle, rounded corners, minimum width=1.5cm, minimum height=0.8cm,text centered, draw=black, fill=green!30]
\tikzstyle{bim} = [rectangle, rounded corners, minimum width=0.1cm, minimum height=0.1cm, text centered, draw=blue, dash pattern=on 4pt off 4pt]
\tikzstyle{arrow} = [thick,->,>=stealth]

\usepackage{makecell}

\newcommand{\beginsupplement}{%
        \setcounter{table}{0}
        \renewcommand{\thetable}{A\arabic{table}}%
        \setcounter{figure}{0}
        \renewcommand{\thefigure}{A\arabic{figure}}%
     }

\newcommand\tf[1]{\textbf{#1}}

\def\ie{\textit{i.e.}}

\newcommand{\myparagraph}[1]{\vspace{1pt}\noindent{\bf{#1}}~~}

\newcommand{\ms}[1]{\tiny{$\pm$#1}}

\usepackage{colortbl}
\definecolor{lightgray}{gray}{0.75}
\definecolor{lightergray}{gray}{0.85}
\definecolor{Blue}{RGB}{3, 31, 97}
\definecolor{Blue1}{RGB}{214, 235, 245}
\definecolor{Blue2}{RGB}{235, 245, 250}
\definecolor{Gray}{RGB}{247, 252, 255}

\definecolor{convcolor}{HTML}{412F8A}
\definecolor{resnetcolor}{HTML}{8DA0CB}
\definecolor{vitcolor}{HTML}{fc8e62}

\begin{document}
\title{Incremental Learning Meets Transfer Learning: Application to Multi-site Prostate MRI Segmentation}
\titlerunning{Incremental Learning Meets Transfer Learning}

\author{Chenyu You\textsuperscript{1(\Letter)}$^\star$, Jinlin Xiang\textsuperscript{2}$^\star$, Kun Su\textsuperscript{2}, Xiaoran Zhang\textsuperscript{3}, Siyuan Dong\textsuperscript{1}, \\ John Onofrey\textsuperscript{4}, Lawrence Staib\textsuperscript{1,3,4},
James S. Duncan\textsuperscript{1,3,4}}

\authorrunning{C. You et al.}

\institute{\textsuperscript{1}Electrical Engineering, Yale University, New Haven, CT USA
\\
\email{chenyu.you@yale.edu}\\
\textsuperscript{2}Electrical and Computer Engineering, The University of Washington, WA USA
\\
\textsuperscript{3}Biomedical Engineering, Yale University, New Haven, CT USA \\
\textsuperscript{4}Radiology \& Biomedical Imaging, Yale School of Medicine, New Haven, CT USA\\
}

\maketitle              
\renewcommand{\thefootnote}{\fnsymbol{footnote}}
\footnotetext[1]{Equal contribution.}
\renewcommand{\thefootnote}{\arabic{footnote}}

\begin{abstract}
Many medical datasets have recently been created for medical image segmentation tasks, and it is natural to question whether we can use them to sequentially train a single model that (1) performs better on all these datasets, and (2) generalizes well and transfers better to the unknown target site domain. Prior works have achieved this goal by jointly training one model on multi-site datasets, which achieve competitive performance on average but such methods rely on the assumption about the availability of all training data, thus limiting its effectiveness in practical deployment. In this paper, we propose a novel multi-site segmentation framework called \textbf{incremental-transfer learning (ITL)}, which learns a model from multi-site datasets in an end-to-end sequential fashion. Specifically, ``incremental" refers to training sequentially constructed datasets, and ``transfer" is achieved by leveraging useful information from the linear combination of embedding features on each dataset. In addition, we introduce our ITL framework, where we train the network including a site-agnostic encoder with pretrained weights and at most two segmentation decoder heads. We also design a novel site-level incremental loss in order to generalize well on the target domain. Second, we show for the first time that leveraging our ITL training scheme is able to alleviate challenging catastrophic forgetting problems in incremental learning. We conduct experiments using five challenging benchmark datasets to validate the effectiveness of our incremental-transfer learning approach. Our approach makes minimal assumptions on computation resources and domain-specific expertise, and hence constitutes a strong starting point in multi-site medical image segmentation.
\keywords{Incremental Learning \and Transfer Learning \and Medical Image Segmentation.}
\end{abstract}

\section{Introduction}
\label{section:intro}

Many medical image datasets have been created over the year, and recent breakthrough achieved by supervised training accelerates the pace in medical image segmentation. Despite great promise, many prior works have limited clinical value, since they are separately trained on small datasets in terms of scale, diversity, and heterogeneity of annotations. As a result, such single-site methods \cite{nie2018asdnet,li2018semi,milletari2016v,jia2019hd,you2020unsupervised,yang2020nuset,you2022simcvd,you2022bootstrapping,yu2017volumetric,zhang2017deep,zhang2018fully,zhang2020fully,zhang2021automatic,zhang2021fully} are vulnerable to unknown target domains, and linearly expand parameters since they assume to train a new model in isolation when adding new datasets. This jeopardizes their trustworthiness and practical deployment in real-world clinical environments.

In this paper, we carry out the \textbf{first-of-its-kind} comprehensive exploration of how to build a multi-site model to achieve strong performance on the training domains and can also serve as a strong starting point for better generalization on new domains in the clinical scenarios. Multi-site training \cite{gibson2018inter,rundo2019use,rundo2020cnn,aslani2020scanner,karani2018lifelong,chang2019domain,dou2020unpaired} has been proposed to consolidate the generalization on multi-site datasets, but it has the following limitations: (1) it still exhibits certain vulnerability to different domains (\ie, different imaging protocols), which yields sub-optimal performance \cite{aslani2020scanner,li2019episodic,you2021momentum}; (2) due to various constraints (\ie, imaging time, privacy, and copyright status), it could become challenging or even infeasible for the requirement on the availability of all training data in a certain time phase. For example, when a new site's data will be available after training, the model requires retraining, which largely prohibits the practical deployments; and (3) consider the relatively small size of the single medical imaging dataset, simply training a dense network from scratch usually leads to sub-optimal segmentation quality because the model might over-fit to those datasets.

Our \textbf{key idea} is to combine the benefits of incremental-learning (IL) and transfer-learning by sequentially training a multi-dataset expert: we continually train a model with corresponding pretrained weights as new site data are incrementally added, which we call \textbf{I}ncremental-\textbf{T}ransfer \textbf{L}earning (ITL). This setting is appealing as: (1) the common IL setting \cite{rebuffi2017icarl,chaudhry2018riemannian,li2017learning,liu2021incremental,wu2019large,davidson2020sequential,zheng2020bi,xiang2022tkil} is to train the base-learner when different site datasets gradually come; thus the effectiveness of this approach heavily depends on the optimality of the base-learner. Consider each single medical image dataset is usually of relatively small size, it is undesirable to build a strong base-learner from scratch; (2) transfer-learning \cite{zhoureview2021,shi2021marginal,yao2021label,zhu2020rubik,you2022class} typically leads to better performance and faster convergence in medical image analysis. Inspired by these findings above, we develop a novel training strategy for expanding its high-quality learning abilities to our multi-site incremental setting, considering both \textit{model-level} and \textit{site-level}. Specifically, our system is built upon a site-agnostic encoder with pretrained weights from natural image datasets such as \textsc{ImageNet}, and at most two segmentation decoder heads wherein only one head is trainable, and the other is fixed associated with specific sites - a parameter-efficient design. Our intuition is that the shared site-agnostic encoder network with pretrained weights encodes regularities across different medical image datasets, while the target and source segmentation decoder heads model the sub-distribution by our proposed site-level incremental loss, resulting in an accurate and robust model that transfers better to new domains without sacrificing performance. We conduct a comprehensive evaluation of ITL on five prostate MRI datasets. Our approach can consistently achieve competitive performance and faster convergence compared to the upper-bound baselines (\ie, isolated-site and mixed-site training), and has a clear advantage on overall segmentation performance compared to the lower-bound baselines (\ie, multi-site training). We also find that our simple approach can effectively address the forgetting issues. Our experiments demonstrate the benefits of modeling both multi-site regularities and site-specific attributes, and thereby serve as a strong starting point on this important practical setting.

\begin{figure}[t]
\centering
\includegraphics[width=\linewidth]{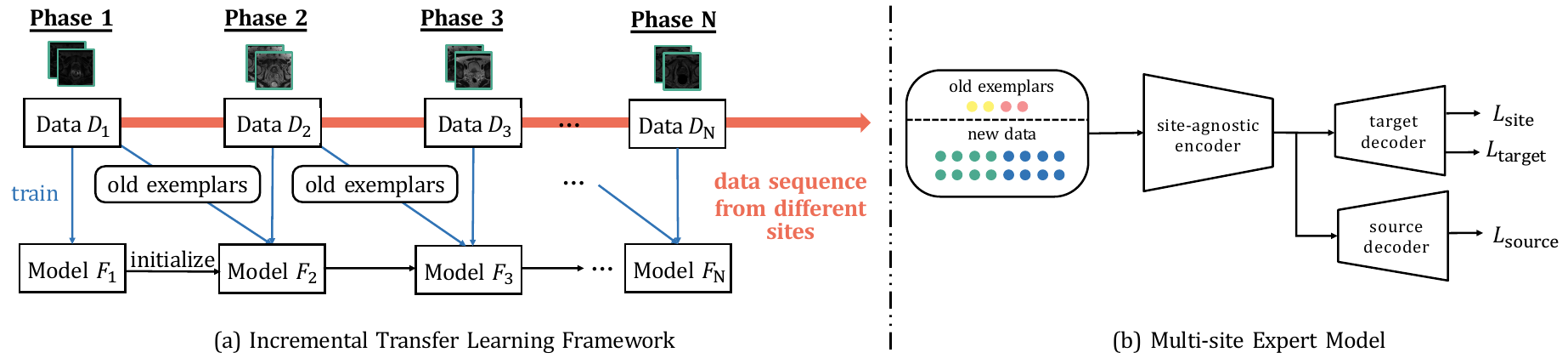}
\vspace{-20pt}
\caption{Overview of (a) our proposed Incremental Transfer Learning framework, and (b) the multi-site expert model. Note that in this study, we only use one multi-site expert model and one source decoder network, which will not introduce additional parameter.} 
\label{fig:model}
\vspace{-10pt}
\end{figure}

\section{Method}
\label{section:method}

\subsection{Problem Setup}
In ITL, a model incrementally learns from a sequential site stream wherein new datasets (namely, medical image segmentation tasks with new sites) are gradually added during the training, as illustrated in Figure \ref{fig:model}. More formally, we denote the sequence of multi-site datasets to be trained as a multi-domain data sequence $\mathcal{D}\!=\!\{D_{1},D_{2},\cdots,D_{N}\}$ of $N$ sites, and $i$-th site $D_{i}$ contains the training images $X\!=\!\{x_j\}_{j=1}^{M}$ and segmentation labels $Y\!=\!\{y_j\}_{j=1}^{M}$, where~${x}_{j} \in \mathbb{R}^{H \times W \times 3}$ is the {augmented} image input, and ${y}_{j} \in \{0, 1\}^{{H \times W}}$ is the ground-truth label. Here the {augmented} input setting is appealing: the axial context naturally provided by a 3D volume can uniquely yield more robust semantic representations to the downstream tasks. We assume access to a multi-site expert model $F_{i}\!=\!\{E_{i},G_{i}\}$ for $i$-th (site) phase, including a pretrained model as a site-agnostic encoder network $E_{i}$ with the weight $\theta_{i}$, a target decoder network $G_{i}^{t}$ with the weight $\theta^{t}_{i}$. During training, we additionally attach a source decoder network $G_{i}^{s}$ (\ie, using $G_{i-1}^{s}$ from previous phrase) with the weight $\theta^{s}_{i}$. In the $i$-th incremental (site) phase, the multi-site expert model has access to two types of domain knowledge: the site-specific knowledge from the current dataset $D_{i}$ and old exemplars $P_{i}$. The latter refers to a set of old exemplars from all previous training datasets $D_{1:i-1}$ in the memory protocol $\mathcal{M}$. This is highly nontrivial to preventing the challenging {``}catastrophic forgetting{"} problem \cite{mccloskey1989catastrophic} of the current dataset $i$ against previous sites in clinical practice. Note that, in this study, we only use one multi-site expert model and one source decoder network, which will not introduce additional parameters. Based on the setting above, we define the ITL problem below.

\myparagraph{Problem of ITL}
\textit{In the current site $i$, our goal is to continuously learn a multi-site expert model based on the knowledge from both $(D_{i},P_{i})$ and the pretrained weight, making the model (1) generalizes well on the unseen data at site $i$, and (2) achieves competitive performance on the previous sites.}

\subsection{Preliminary}
Our goal is to build a strong multi-site model by learning a site-agnostic encoder with pretrained weights as well as a segmentation decoder over multi-site datasets. This naturally raises several interesting questions: \textit{How well will ITL-based methods perform in multi-site medical image datasets? Will transfer learning make the base learner stronger on the unseen site? If yes, can they perform stably well?} To answer the above questions, a prerequisite is to define the upper bound and lower bound. Here we introduce three common paradigms for multi-site medical image segmentation: (1) isolated-site training, (2) mixed-site training, and (3) multi-site training. It is well-known that the isolated-site and mixed-site training approaches can achieve state-of-the-art performance when evaluating the same dataset, while the performance catastrophically drops when evaluating new datasets. On the other hand, the multi-site training approach often yields inconsistent performance across multiple sites. For all training paradigms, we minimize Dice loss between the predicted outputs and the ground truth label.

\begin{table}[t]
\centering
\caption{
Information about five different sites from three benchmark datasets.
}
\vspace{-5pt}
\label{tab:datasets}
\resizebox{0.80\linewidth}{!}{%
\begin{tabular}{c|c|c|c|c|c|c}
\toprule
\textbf{Dataset} & \tf{Modality}  & \tf{\# of cases} & \tf{Field
strength (T)} & \tf{Resolution(in/through plane)(mm)} &\tf{Coil} & \tf{Source}\\
\midrule
Site0 & MRI & 30 & 3 & 0.6-0.625\,/\,3.6-4 & Surface & NCI-ISBI13 \cite{bloch2015nci} \\
Site1 & MRI & 30 & 1.5 & 0.4\,/\,3.0 & Endorectal & NCI-ISBI13 \cite{bloch2015nci} \\
Site2 & MRI  & 19 & 3 & 0.67-0.79\,/\,1.25 & - & I2CVB \cite{lemaitre2015computer}\\
Site3 & MRI & 12 & 1.5 & 0.625\,/\,3.6 & Endorectal & PROMISE12 \cite{litjens2014evaluation}\\
Site4 & MRI & 13 & 1.5 and 3 & 0.325-0.625\,/\,3-3.6 & - & PROMISE12 \cite{lemaitre2015computer} \\
\bottomrule
\end{tabular}
}
\vspace{-10pt}
\end{table}

\myparagraph{Upper Bound}
We consider two training paradigms (\ie, isolated-site and mixed-site training) as our upper bound baseline. For isolated-site training, given each site $D_{i}$, we train isolated-site models separately. The architecture of the isolated-site model consists of a pretrained encoder $E_{i}$ and a segmentation decoder network, same architecture as $G_{i}$. Then, we apply different isolated-site models to predict results based on the site-specific data at inference. However, this approach dramatically increases memory and computational overhead, making it practically challenging at scale. For mixed-site training, we train one full model on the full mixed-site data $D$, and then use the well-trained model for inference. However, it requires the simultaneous presence of all data in training and inference.

\myparagraph{Lower Bound}
For multi-site training, we sequentially train only one model coupled with the pretrained weights on all sites. This can get rid of large parameter counts, making it appealing in practice. However, due to the forgetting quandary, it inevitably suffers from severe performance degradation. This naturally questions: \textit{can we improve performance on multi-site medical image segmentation with minimal additional memory footprint?} In the following, we give an affirmative answer.

\subsection{Proposed Incremental Transfer Learning Multi-Site Method}
To address the aforementioned problems, we develop the incremental transfer learning framework to perform well on the training distribution and generalize well on the new site dataset with minimal additional memory. To our best knowledge, we are \tf{the first work} to apply incremental transfer learning to the limited clinical data regimes. To control the parameter efficiency, we decompose the model into a share site-agnostic encoder $E_{i}$ and two segmentation decoder heads (\ie, source decoder $G_{i}^{s}$ and target decoder $G_{i}^{t}$). In this way, we can keep the network parameters the same when adding a new site. 
Specifically, $G_{i}^{s}$ is designed to transfer the knowledge of a previously learned site, and $G_{i}^{t}$ is designed to comprehensively train on a new site and previous datasets. During training, we only update $G_{i}^{t}$ while $G_{i}^{s}$ is frozen. It is worth mentioning that our proposed framework is independent of the encoder architecture, and can be easily plugged in other pretrained vision models.

The full ITL algorithm is summarized in Algorithm \ref{tab:algorithm}. We describe our ITL algorithm as follows. We first randomly initialize $G_{i}^{t}$, $G_{i}^{s}$, and then iteratively train our full model (\ie, a pretrained encoder $E_{i}$ and two decoders $G_{i}^{t}$, $G_{i}^{s}$) with $N$-site training samples. Bounded by the computational requirements, it is challenging or even infeasible to retain all data for training. Inspired by recent work \cite{rebuffi2017icarl}, to maintain the knowledge of previous sites, we {``}store{"} all the old site data exemplars in the memory protocol $\mathcal{M}_{i}$. In the $i$-th incremental (site) phase, we first load $P_{i}$, and then use both $P_{i}$ and $D_i$ to train $F_{i}$ initialized by $\theta_{i}^{s}$. This setting is appealing as (1) it can substantially alleviate the imbalance between the old and new site knowledge, and (2) it is efficient to train on them. Of note, we do not use the source decoder when training on the first-site dataset. We formulate ITL as \textit{model-level} and \textit{site-level} optimization. 

\myparagraph{Model-level Optimization}
To perform better on all these training distributions, we propose improving generic representations by distilling knowledge from previous data. In each incremental phase, we jointly optimize two groups of learnable parameters in our ITL learning by minimizing the \textit{model-level} incremental loss (\ie, $\mathcal{L}_{\text{model}}\!=\!\mathcal{L}_{\text{target}}+\mathcal{L}_{\text{source}}$) on all training samples (\ie, $D_{i}\bigcup D_{0:i-1}$): (1) a share site-agnostic encoder $E_{i}$ and a target decoder $G_{i}^{t}$; (2) a share site-agnostic encoder $E_{i}$ and a source decoder $G_{i}^{s}$. This helps ITL avoid catastrophic forgetting of prior site-specific knowledge.

\begin{algorithm}[t]
\caption{Incremental-Transfer Learning(ITL) Algorithm}
\label{tab:algorithm}
\begin{algorithmic}[1]
    \REQUIRE Dataset: $\mathcal{D}$; Hyper-parameters: $\alpha, \delta, \gamma$
    \STATE Initialize the $\mathcal{M}\textit{ (Memory)}$ : $\mathcal{M}$
    \STATE Initialize the Model $F_0$: $\textit{Pertrained Encoder} \longrightarrow E_0, G_0$
    
        \FOR {$i$ = 1,2,3,....N }
            \FOR {All training Sample in $D_i$ and $\mathcal{M}_{i-1}$}
                \STATE $\mathcal{L}_{\text{target}} = \sum_{j = 0}^{N-1} \alpha_j \mathcal{L}_{\text{Dice}}^{E_i,G_i^{t}}(\mathcal{M}_{j}, Y_j) $ or $0$ When $N = 1$
                \STATE $\mathcal{L}_{\text{source}} = \sum_{j = 0}^{N-1} \delta_j \mathcal{L}_{\text{Dice}}^{E_i,G_{i}^{s}}(\mathcal{M}_{j}, Y_j)$ or $0$ When $N = 1$
                \STATE $\mathcal{L}_{\text{model}}  = \mathcal{L}_{\text{source}} + \mathcal{L}_{\text{target}}$
                 \STATE $\mathcal{L}_{\text{site}} = \mathcal{L}_{\text{Dice}}^{E_i,G_{i}^{t}}(\mathcal{D}_i, Y_i)$
                \STATE $\mathcal{L}_{\text{all}} = \mathcal{L}_{\text{site}} + \mathcal{L}_{\text{model}}  $
                \STATE $F_i = (E_i, G_{i}^{t})$ by minimizing the $\mathcal{L}_{\text{all}}$
            
            \ENDFOR
            \STATE Update Memory: $ \mathcal{M} + \gamma \% D_N \longrightarrow \mathcal{M}$
            \STATE Save Teacher Model: $G_N$
        \ENDFOR
\end{algorithmic}
\end{algorithm}

\myparagraph{Site-level Optimization}
The above \textit{model-level} optimization is used to maintain previously learned knowledge. In contrast, this step is design to train the  multi-site model to learn site-specific knowledge on the newly added site. Specifically, we minimize the site-level incremental loss $\mathcal{L}_{\text{site}}$ between the probability distribution from $F_{i}$ and the ground truth. This essentially learns the site-specific knowledge for the downstream medical image segmentation tasks. Of note, $\mathcal{L}_{\text{source}}$, $\mathcal{L}_{\text{target}}$, and $\mathcal{L}_{\text{site}}$ use the Dice loss. The overall loss combines the \textit{model-level} loss and the \textit{site-level} loss as follows:
\begin{equation}
\mathcal{L}_{\text{all}} = \mathcal{L}_{\text{model}}+\mathcal{L}_{\text{site}}.
\label{equation:all}
\end{equation}

\section{Experiments}
\label{section:exp}
\myparagraph{Datasets and Settings}
We evaluate our proposed incremental transfer learning method on three prostate T2-weighted MRI datasets with different sub-distributions: NCI-ISBI13 \cite{bloch2015nci}, I2CVB \cite{lemaitre2015computer}, and PROMISE12 \cite{litjens2014evaluation}. Due to the diverse data source distributions, they can be split into five multi-site datasets, which is similar to \cite{liu2020ms}. Table \ref{tab:datasets} provides some dataset statistics. For pre-processing, we follow the setting in \cite{liu2020saml} to normalize the intensity, and resample all 2D slices and the corresponding segmentation maps to $384\times 384$ in the axial plane. For all five site datasets, we randomly split each original site dataset into training and testing with a ratio of 4 : 1. For each site training, we divide the data from the previous site into a small subset with a certain portion (\ie,  1\%, 3\%, 5\%), and combine it with the current site data for training.

\begin{table*}[t]
	\begin{center}
	\caption{Comparison of segmentation performance (DSC{[}\%{]}/95HD{[}mm{]}) across datasets. Note that a larger DSC ($\uparrow$) and a smaller 95HD ($\downarrow$) indicate better performing ITL models. We use four models pretrained on \textsc{ImageNet}: ResNet-18, ResNet-34, ResNet-50, and ViT under different portions (\ie, 1\%, 3\%, 5\%) of exemplars from previous data for every incremental phase. We consider multi-site training as the lower bound, isolated-site, and mixed-site training as the upper bound.}
	\vspace{-5pt}
	\label{tab:site_main}
    \begin{adjustbox}{width=0.8\linewidth}
	\begin{tabular}{cccccccccccc}\toprule
		 & &
		 \multicolumn{2}{c}{HK} &
		 \multicolumn{2}{c}{UCL} &
		 \multicolumn{2}{c}{ISBI} &
		 \multicolumn{2}{c}{ISBI1.5 } &
		 \multicolumn{2}{c}{I2CVB}
		 \\
         \cmidrule(r){3-4} \cmidrule(r){5-6} \cmidrule(r){7-8} \cmidrule(r){9-10} \cmidrule(r){11-12}
         {Backbone}
         & {Scheme} &DSC{[}\%{]}&95HD{[}mm{]}&DSC{[}\%{]}&95HD{[}mm{]}&DSC{[}\%{]}&95HD{[}mm{]}&DSC{[}\%{]}&95HD{[}mm{]}&DSC{[}\%{]}&95HD{[}mm{]}\\
         \midrule
		\multirow{6}{*}{RES-18}
                    & Multi
		            & {59.38}
		            &{64.17} 
		            & {66.26}
		            &{54.19}
                    & {54.38}
                    &{73.40}
                    & {66.89}
                    & {44.49}
		            & {84.54}
		            & {11.70} 
                    \\
                    &1\% 
		            & {67.82}
		            &{56.08} 
		            & {67.12}
		            &{58.05}
                    & {59.47}
                    &{70.46}
                    & {77.34}
                    & {34.77}
		            & {82.94}
		            &{ 6.06} 
                    \\
                    &3\% 
		            & {71.60}
		            &{18.41} 
		            & {82.18}
		            &{23.92}
                    & {72.26}
                    & {20.91}
                    & {81.53}
                    & {19.21}
		            & {84.08}
		            & {13.75} 
                    \\
                    &5\% 
		            & {81.81}
		            &{5.50} 
		            & {84.45}
		            &{13.95}
                    & {84.52}
                    & {15.65}
                    & {89.32}
                    & {10.11}
		            & {86.72}
		            & {11.70} 
                    \\
                    & Isolated 

		            & {93.46}
		            &{2.06} 
		            & {88.29}
		            &{6.20}
                    & {93.35}
                    &{2.04}
                    & {90.89}
                    & {7.53}
		            & {88.74}
		            &{ 13.93} 
                    \\
                    & Mixed
		            & {92.17}
		            &{7.60} 
		            & {83.38}
		            &{12.22}
                    & {91.70}
                    &{2.46}
                    & {90.08}
                    &{9.20}
		            & {89.12}
		            &{ 13.86} 
                    \\\midrule 
        \multirow{6}{*}{RES-34} 
                    &{Multi}
                    & {57.75}
                    &{55.13}
		            & {64.87}
		            & {52.50}
		             & {57.47}
		            &{65.38} 
		            & {65.61}
		            &{56.83}
                    & {91.46}
                    &{8.83}
                    \\
                    &  1\% 
                    & {67.40}
                    & {24.18}
		            & {79.55}
		            &{ 30.43}
		            & {69.61}
		            & {44.69}
		            & {84.68}
		            &{18.71}
                    & {89.38}
                    & {15.24}
                    \\
                    &  3\% 
                    & {80.90}
                    & {28.41}
		            & {82.57}
		            &{ 22.18}
		            & {75.89}
		            &{26.26} 
		            & {84.68}
		            &{10.57}
                    & {90.35}
                    &{13.15}
                    \\
                    &  5\% 
                    & {80.46}
                    &{22.92}
		            & {87.79}
		            & {17.32}
		            & {88.14}
		            & {14.64} 
		            & {90.29}
		            & {8.57}
                    & {91.30}
                    & {8.52}
                    \\
                    & Isolated
                    &{93.87}
                    &{1.89}
                    &{89.03}
                    &{4.05}
                    &{92.08}
                    &{2.19}
                    &{92.57}
                    &{7.96}
                    &{91.57}
                    &{7.98}
                    \\
                    & Mixed
                    & {93.85}
                    &{1.71}
		            & {87.81}
		            &{ 16.85}
		            & {91.49}
		            &{3.35} 
		            & {93.82}
		            &{5.30}
                    & {92.58}
                    &{6.64}
                    \\
                    \midrule 
        \multirow{6}{*}{RES-50}  
                    & Multi
                    & {63.24}
                    & {53.98}
		            & {64.79}
		            & {56.59}
		            & {72.95}
		            & {26.63} 
		            & {69.41}
		            & {49.89}
                    & {90.40}
                    & {8.21}
                    \\
                    &  1\% 
                    & {69.01}
                    & {60.70}
		            & {69.85}
		            & {44.21}
		            & {75.30}
		            & {28.74} 
		            & {80.27}
		            & {20.10}
                    & {90.08}
                    & {8.02}
                    \\
                    &  3\% 
                    & {78.72}
                    & {16.89}
		            & {83.74}
		            & {12.81}
		            & {84.96}
		            & {8.51} 
		            & {86.95}
		            & {6.18}
                    & {92.34}
                    & {5.24}
                    \\
                    &  5\% 
                    & {92.46}
                    & {2.92}
		            & {88.79}
		            & {10.97}
		            & {92.16}
		            & {2.04} 
		            & {92.18}
		            & {4.87}
                    & {91.35}
                    & {2.12}
                    \\
                    & Isolated
                    &{93.73}
                    &{2.12}
                    &{89.03}
                    &{7.23}
                    &{93.26}
                    &{4.39}
                    &{93.48}
                    &{5.10}
                    &{93.20}
                    &{2.40}
                    \\
                    & Mixed
                    & {94.38}
                    & {1.34}
		            & {88.28}
		            & {9.77}
		            & {92.71}
		            & {9.43} 
		            & {92.27}
		            &{5.29}
                    & {90.45}
                    &{5.29}
                    \\
                    \midrule 
        \multirow{6}{*}{VIT} 
                    & Multi
		            & {66.94}
		            & {53.57} 
		            & {65.85}
		            & {54.69}
                    & {92.66}
                    & {6.37}
                    & {72.80}
                    & {51.35}
		            & {90.56}
		            & {7.02} 
                    \\
                    & 1\% 
		            & {71.99}
		            &{48.61} 
		            & {85.29}
		            &{11.35}
                    & {75.99}
                    & {17.87}
                    & {84.73}
                    & {12.32}
		            & {90.11}
		             & {7.23}
                    \\
                    & 3\% 
		            & {79.33}
		            & {20.84} 
		            & {88.16}
		            &{7.08}
                    & {85.48}
                    & {7.97}
                    & {87.64}
                    & {9.95}
		            & {90.07}
		            &{ 6.94} 
                    \\
                    &  5\% 
		            & {93.25}
		            & {1.37} 
		            & {87.62}
		            & {9.23}
                    & {92.22}
                    & {4.82}
                    & {91.62}
                    & {2.82}
		            & {91.87}
		            & {6.59} 
                    \\
                    & Isolated
		            & {94.44}
		            &{1.88} 
		            & {88.80}
		            &{8.21}
                    & {93.23}
                    &{4.76}
                    & {92.47}
                    & {6.27}
		            & {93.23}
		            &{ 6.43} 
                    \\
                    & Mixed
		            & {93.30}
		            & {1.38} 
		            & {87.20}
		            & {9.21}
                    & {92.86}
                    & {9.29}
		            & {86.92}
		            & {12.28} 
		            & {92.01}
                    & {6.99}
                    \\           \bottomrule
	\end{tabular}
    \end{adjustbox}
    \end{center}
    \vspace{-20pt}
\end{table*}

\myparagraph{Training and Evaluation}
In this study, we implement all models using Pytorch. We set $H,W$ as 384, $\alpha,\delta$ as 0.5, and the batch size as $5$. To mitigate the overfitting, we augment the data by random horizontal flipping, random rotation, and random shift. We adopt ResNet family \cite{he2016deep} (\ie, ResNet18, ResNet34, ResNet50) and ViT \cite{dosovitskiy2020image} (\ie, R50+ViT-B/16 hybrid model) as our pretrained encoder. We evaluate the model performance by Dice coefficient (DSC) and 95\% Hausdorff Distance (95HD). For a fair comparison, we adopt the same decoder architecture design in \cite{liu2020saml} are shown in Appendix Table \ref{tab:decoder}, and do not use any post-processing techniques. All of our experiments are conducted on two NVIDIA Titan X GPUs. All the models are trained using Adam optimizer with $\beta_1=0.9$, $\beta_2 = 0.999$. For 100 epochs training, a multi-step learning rate schedule is initialized as $0.001$ and then decayed with a power of $0.95$ at epochs $60$ and $80$.

\myparagraph{Main Results}
We conduct extensive experiments on five benchmark datasets. We adopt four models: ResNet-18, ResNet-34, ResNet-50, and ViT. We select three portions (\ie, 1\%, 3\%, 5\%) of exemplars from previous data for every incremental phase. Our results are presented in Table \ref{tab:site_main} and Appendix Figure \ref{fig:vis_multi}. First and foremost, we can see ITL-based methods generalize across all datasets under two exemplar portions (\ie, 3\% and 5\%), yielding the competitive segmentation quality comparable to the upper bound baselines (\ie, isolated-site and mixed-site training), which are much higher than the lower bound counterparts. The 1\% exemplar portion seems slightly more challenging for ITL, but its superiority over the lower bound counterparts is still solid. A possible explanation for this finding is that using two exemplar portions (\ie, 3\% and 5\%) maintains enough information of ITL, which mitigates the catastrophic forgetting, while ITL trained in the setting of 1\% exemplar portion is not powerful enough to inherit prior knowledge and generalize well on newly added sites. Second, we consistently observe that ITL using larger models (\ie, ResNet-50 and ViT) generalize substantially better than those using small models (\ie, ResNet-18 and ResNet-34), which demonstrate competitive performance across all datasets. These results suggest that our ITL using the large model as our pretrained encoder leads to substantial gains in the setting of very limited data.
\section{Analysis and Discussion}
We address several research questions pertaining to our ITL approach. We use a ResNet-18 model as our encoder in our experiments. For comparisons, all models are trained for the same number of epochs, and all results are the average of three independent runs of experiments. To study the effectiveness of our proposed ITL framework, we performed experiments with $5\%$ exemplars ratio.

\myparagraph{Does transfer learning lead to better ITL?}
We draw two perspectives that may intuitively explain the effectiveness of transfer learning in our proposed ITL framework. As a first test of \textit{whether transfer learning makes the base-learner stronger}, we plot the training loss/validation loss (\ie, $\mathcal{L}_{\text{all}}$) to iteration to demonstrate the convergence improvements in Appendix Figure \ref{fig:ablation1}. We can see that training from pretrained weights can converge faster than training from scratch. Another (perhaps not so surprising) observation we can get from Appendix Figure \ref{fig:ablation1} is that using pretrained weights usually yields slightly smaller loss compared to training from scratch. We then ask \textit{whether transfer learning produces increased performance on multi-site datasets}. Since each single medical image dataset is usually of relatively small size, training the model from scratch tends to overfit a particular dataset. To evaluate the impact of transferring learning, we compare w/pretraining to w/o pretraining. As shown in Appendix Table \ref{tab:component}, training from scratch does not bring benefits to the ITL framework. Instead of training from scratch, we find that simply incorporating transfer learning significantly boots the performance of ITL while achieving faster convergence speed, suggesting that transfer learning provides additional regularization against overfitting.

\begin{table*}[t]
	\begin{center}
	\caption{Comparison of segmentation performance in different phases.}
	\vspace{-5pt}
	\label{tab:incremental}
    \begin{adjustbox}{width=0.7\linewidth}
	\begin{tabular}{cccccccccc}\toprule
		 
		 \multicolumn{2}{c}{HK} &
		 \multicolumn{2}{c}{UCL} &
		 \multicolumn{2}{c}{ISBI} &
		 \multicolumn{2}{c}{ISBI1.5 } &
		 \multicolumn{2}{c}{I2CVB}
		 \\
        \cmidrule(r){1-2} \cmidrule(r){3-4} \cmidrule(r){5-6} \cmidrule(r){7-8} \cmidrule(r){9-10}
          DSC{[}\%{]}&95HD{[}mm{]}&DSC{[}\%{]}&95HD{[}mm{]}&DSC{[}\%{]}&95HD{[}mm{]}&DSC{[}\%{]}&95HD{[}mm{]}&DSC{[}\%{]}&95HD{[}mm{]}\\
         \midrule

		            {94.06}
		            &{1.96} 
		            & -
		            &-
                    &-
                    &-
                    & -
                    & -
		            & -
		            & -
                    \\
  
		            {93.68}
		            &{1.98} 
		            & {88.74}
		            &{8.72}
                    & -
                    &-
                    & -
                    & -
		            &-
		            &-
                    \\
         
		            {93.20}
		            &{1.83} 
		            & {87.38}
		            & {9.30}
                    & {92.87}
                    & {1.82}
                    & -
                    & -
		            & -
		            & -
		                                \\  
		              {90.37}
		            &{8.34} 
		            & {86.73}
		            & {12.75}
                    & {89.84}
                    & {13.32}
                    & {91.57}
                    & {11.05}
		            & -
		            & -
		                                \\  
		          {88.88}
		            &{8.91} 
		            & {85.14}
		            & {10.97}
                    & {85.98}
                    & {14.23}
                    & {89.74}
                    & {13.11}
		            & {88.46}
		            & {12.15}
		                                \\                        \bottomrule
	\end{tabular}
    \end{adjustbox}
    \end{center}
    \vspace{-20pt}
\end{table*}

\myparagraph{Does ITL generalizes well on multi-site datasets?}
We investigate whether the ITL framework generalizes well on multi-site datasets. We report the segmentation results of different phases in Table \ref{tab:incremental}, from which we observe that ITL achieves good performance in different phases. This reveals that our approach is greatly helpful in reducing forgetting issues. We evaluate the proposed ITL methods with two random ordering (\ie, (1) \{HK$\rightarrow$UCL$\rightarrow$ISBI$\rightarrow$ISBI1.5$\rightarrow$I2CVB\}, and (2) \{ISBI$\rightarrow$ISBI1.5$\rightarrow$I2CVB$\rightarrow$HK$\rightarrow$ UCL\}). The results are shown in Appendix Table \ref{tab:sequence}. We perform experiments using both ordering strategies and observe comparable performance.

\myparagraph{Efficiency of ITL}
We report the network size and memory costs in Appendix Table \ref{tab:cost}. We observe that ITL achieves competitive performance and utilizes less network parameters compared to isolated-site training (upper bound), which requires the new model when adding new site data. We also examine the required memory footprint at each incremental phase. We observe that ITL is significantly more memory-efficient than mixed-site training (upper bound), although the latter remains the same network size when adding a new training phase. These results further demonstrate the efficiency of our proposed ITL framework.
\section{Conclusion}
\label{section:conclusion}
In this paper, we present a novel incremental transfer learning framework for incrementally tackling multi-site medical image segmentation tasks. We pose model-level and site-level incremental training strategies for better segmentation, generalization, and transfer performance, especially in limited clinical resource settings. Extensive experimental results on four different baseline architectures demonstrate the effectiveness of our approach, offering a strong starting point to encourage future work in these important practical clinical scenarios.

%
%
%
\bibliographystyle{splncs04}
\bibliography{ref}

\clearpage
\setcounter{page}{1}

\appendix
\section*{Appendix}
\beginsupplement

\begin{figure}[ht!]
\centering
\includegraphics[width=\linewidth]{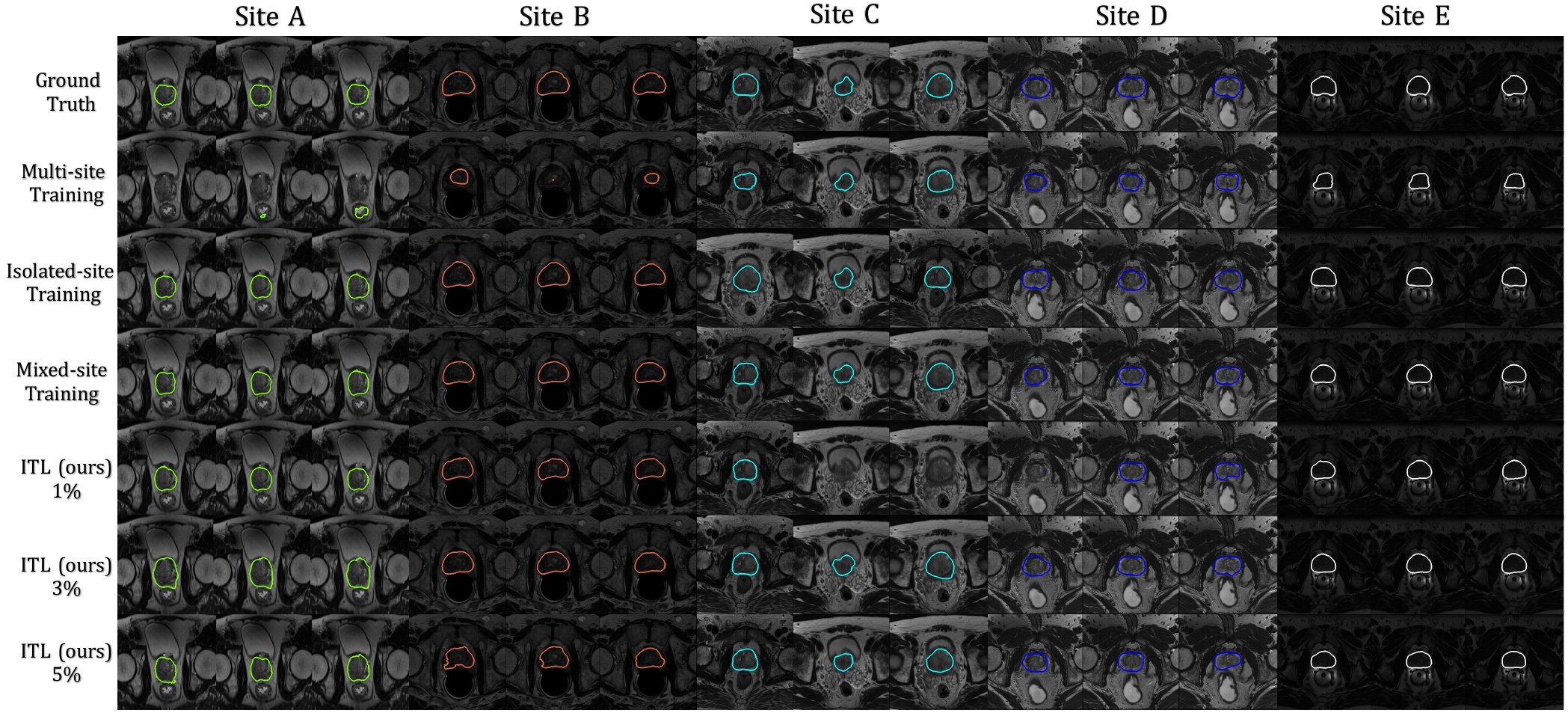}
\vspace{-10pt}
\caption{Visualization of segmentation results on five benchmarks using ResNet-18 as the encoder. Different site results are shown in different colors.} 
\label{fig:vis_multi}
\vspace{-15pt}
\end{figure}

\begin{table}[ht]
\begin{center}
\caption{Segmentation decoder head architecture}
\vspace{-5pt}
\label{tab:decoder}
\begin{adjustbox}{width=0.3\linewidth}
\begin{tabular}{cc} 
\toprule
\multicolumn{2}{c}{\tf{Deocder}}     \\ 
\midrule
Layer            & Feature size \\
\midrule
Upsample 1          & 48x48        \\
Residual block 1    & 48x48        \\
Upsample 2          & 96x96        \\
Residual block 2    & 96x96        \\
Upsample 3          & 192x192      \\
Residual block 3    & 192x192      \\
Upsample 4          & 384x384      \\
Residual block 4    & 384x384      \\
Output prediction   & 384x384      \\
\bottomrule
\end{tabular}
\end{adjustbox}
\end{center}
\end{table}

\begin{table}[ht]
\centering
\caption{Comparison of different ordering strategies using ResNet-18. We report mean and standard deviation across three random trials. Note that a larger DSC ($\uparrow$) and a smaller 95HD ($\downarrow$) indicate better performing ITL models.}
\vspace{-5pt}
\label{tab:sequence}
\begin{adjustbox}{width=0.6\linewidth}
\begin{tabular}{ccc}
\toprule 
{Training sequence} 
& DSC{[}\%{]} 
& 95HD{[}mm{]} \\ 
\midrule
HK$\rightarrow$UCL$\rightarrow$ISBI$\rightarrow$ISBI1.5$\rightarrow$I2CVB 
& 85.36\ms{0.33}
& 11.38\ms{0.36}  
\\
ISBI$\rightarrow$ISBI1.5$\rightarrow$I2CVB$\rightarrow$HK$\rightarrow$UCL 
& 86.27\ms{0.27}
& 12.01\ms{0.68}  
\\ 
\bottomrule
\end{tabular}
\end{adjustbox}
\vspace{-10pt}
\end{table}
\begin{table}[htpb]
\centering
\caption{Comparison of different training strategies using ResNet-18. We report mean and standard deviation across three random trials.}
\vspace{-5pt}
\label{tab:cost}
\resizebox{0.9\linewidth}{!}{
\begin{tabular}{cccccc} 
\toprule 
{Scheme} 
& DSC{[}\%{]} 
& 95HD{[}mm{]}
& {Model size(Mb)}         
& {Add new sites?}   
& {Memory Cost}       
\\ 
\midrule
Isolated 
& 90.95\ms{0.27}
& 6.35\ms{0.68}
& 77.9$\times$Site Num.       
& Linearly increase 
& new Data         
\\
Mixed 
& 89.29\ms{0.38}
& 9.06\ms{0.84}
& 77.9       
& {constant} 
& old Data + new Data         
\\
ITL 
& 85.36\ms{0.33}
& 11.38\ms{0.36}
& 77.9
& {constant} 
& {$5\%$ old Data + new Data }
\\ 
\bottomrule
\end{tabular}}
\vspace{-10pt}
\end{table}
\begin{table*}[ht]
	\begin{center}
	\caption{Ablation of each component in the proposed ITL when using ResNet-18 under 5\% exemplar portion. We report mean and standard deviation across three random trials. Note that a larger DSC ($\uparrow$) and a smaller 95HD ($\downarrow$) indicate better performing ITL models. The best results are in \tf{bold}.}
    \vspace{-5pt}
	\label{tab:component}
    \begin{adjustbox}{width=\linewidth}
	\begin{tabular}{cccccccccccccc}\toprule
		 & &
		 \multicolumn{2}{c}{HK} &
		 \multicolumn{2}{c}{UCL} &
		 \multicolumn{2}{c}{ISBI} &
		 \multicolumn{2}{c}{ISBI1.5 } &
		 \multicolumn{2}{c}{I2CVB}& &
		 \\
         \cmidrule(r){3-4} \cmidrule(r){5-6} \cmidrule(r){7-8} \cmidrule(r){9-10} \cmidrule(r){11-12}
         \textbf{Backbone}
         & \textbf{Component} &DSC{[}\%{]}&95HD{[}mm{]}&DSC{[}\%{]}&95HD{[}mm{]}&DSC{[}\%{]}&95HD{[}mm{]}&DSC{[}\%{]}&95HD{[}mm{]}&DSC{[}\%{]}&95HD{[}mm{]} 	 & Avg. DSC & Avg. 95HD \\
         \midrule
		\multirow{3}{*}{RES-18}
                    & pretraining only
		            & {59.38}
		            &{64.17} 
		            & {66.26}
		            &{54.19}
                    & {54.38}
                    &{73.40}
                    & {66.89}
                    &{44.49}
		            & {81.54}
		            & {28.70} 
		            & {65.69}\ms{1.51}
		            & {52.99}\ms{0.72}
                    \\
                    & $\mathcal{L}_{\text{model}}$ only
		            & {74.02}
		            &{18.86} 
		            & {73.79}
		            &{31.90}
                    & {51.23}
                    &{51.66}
                    & {80.89}
                    &{21.96}
		            & {80.88}
		            &{ 58.44}
		            & {72.16}\ms{0.38}
		            & {36.56}\ms{0.85}
                    \\
                    & pretraining + $\mathcal{L}_{\text{model}}$
		            & \tf{81.81}
		            & \tf{5.50} 
		            & \tf{84.45}
		            & \tf{13.95}
                    & \tf{84.52}
                    & \tf{15.65}
                    & \tf{89.32}
                    & \tf{10.11}
		            & \tf{86.72}
		            & \tf{11.70} 
		            & \tf{85.36}\ms{0.33}
		            & \tf{11.38}\ms{0.36}
                    \\
     \bottomrule
	\end{tabular}
    \end{adjustbox}
    \end{center}
    \vspace{-20pt}
\end{table*}

\begin{figure}[ht]
\centering
\includegraphics[width=\linewidth]{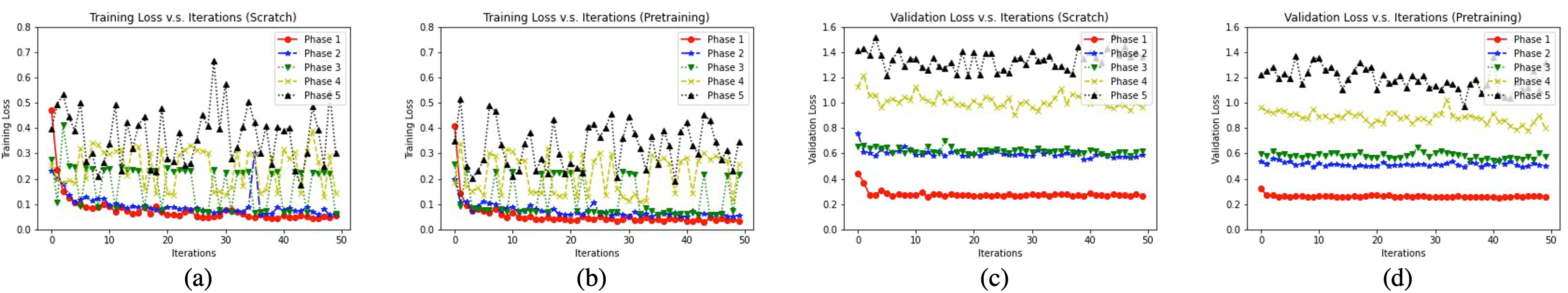}
\vspace{-10pt}
\caption{Comparison of training from scratch against using pretraining. We use ResNet-18 on \textsc{ImageNet} as the encoder. Under 5\% exemplar portion, we plot (a) training loss (Scratch), (b) training loss (Pretraining), (c) validation loss (Scratch), (d) validation loss (Pretraining).} 
\label{fig:ablation1}
\end{figure}

\end{document}